\documentclass[10pt,twocolumn,letterpaper]{article}

\usepackage{cvpr}
\usepackage{times}
\usepackage{epsfig}
\usepackage{graphicx}
\usepackage{amsmath}
\usepackage{amssymb}
\usepackage{subcaption}
\usepackage{multirow}
\usepackage[space]{grffile}
\usepackage{array}
\newcolumntype{C}[1]{>{\centering\arraybackslash}p{#1}}
\usepackage{chngpage}

\newcolumntype{H}{>{\setbox0=\hbox\bgroup}c<{\egroup}@{}}

\usepackage[pagebackref=true,breaklinks=true,letterpaper=true,colorlinks,bookmarks=false]{hyperref}
\usepackage[font=small,labelfont=bf,tableposition=top]{caption}

\cvprfinalcopy % *** Uncomment this line for the final submission

\def\cvprPaperID{1330} % *** Enter the CVPR Paper ID here

\ifcvprfinal\fi
\begin{document}

%%%%%%%%% TITLE
\title{Accurate Image Super-Resolution Using Very Deep Convolutional Networks}

\author{Jiwon Kim, Jung Kwon Lee and Kyoung Mu Lee\\
	Department of ECE, ASRI, Seoul National University, Korea\\
	{\tt\small \{j.kim, deruci, kyoungmu\}@snu.ac.kr}
}

\maketitle
%\thispagestyle{empty}

%%%%%%%%% ABSTRACT
\begin{abstract}
We present a highly accurate single-image super-resolution (SR) method. Our method uses a very deep convolutional network inspired by VGG-net used for ImageNet classification \cite{simonyan2015very}. We find increasing our network depth shows a significant improvement in accuracy. Our final model uses 20 weight layers. By cascading small filters many times in a deep network structure, contextual information over large image regions is exploited in an efficient way. With very deep networks, however, convergence speed becomes a critical issue during training. We propose a simple yet effective training procedure. We learn residuals only and use extremely high learning rates ($10^4$ times higher than SRCNN \cite{dong2015image}) enabled by adjustable gradient clipping. Our proposed method performs better than existing methods in accuracy and visual improvements in our results are easily noticeable.
\end{abstract}

%%%%%%%%% BODY TEXT
\section{Introduction}
We address the problem of generating a high-resolution (HR) image given a low-resolution (LR) image, commonly referred as single image super-resolution (SISR) \cite{Irani1991}, \cite{freeman2000learning}, \cite{glasner2009super}. SISR is widely used in computer vision applications ranging from security and surveillance imaging to medical imaging where more image details are required on demand.

Many SISR methods have been studied in the computer vision community. Early methods include interpolation such as bicubic interpolation and Lanczos resampling \cite{duchon1979lanczos} more powerful methods utilizing statistical image priors \cite{sun2008image,Kim2010} or internal patch recurrence \cite{glasner2009super}.

Currently, learning methods are widely used to model a mapping from LR to HR patches. Neighbor embedding \cite{chang2004super,bevilacqua2012} methods interpolate the patch subspace. Sparse coding \cite{yang2010image,zeyde2012single,Timofte2013,Timofte} methods use a learned compact dictionary based on sparse signal representation. Lately, random forest \cite{schulter2015fast} and convolutional neural network (CNN) \cite{dong2015image} have also been used with large improvements in accuracy.

Among them, Dong et al. \cite{dong2015image} has demonstrated that a CNN can be used to learn a mapping from LR to HR in an end-to-end manner. Their method, termed SRCNN, does not require any engineered features that are typically necessary in other methods \cite{yang2010image,zeyde2012single,Timofte2013,Timofte} and shows the state-of-the-art performance.

While SRCNN successfully introduced a deep learning technique into the super-resolution (SR) problem, we find its limitations in three aspects: first, it relies on the context of small image regions; second, training converges too slowly; third, the network only works for a single scale.

%%%%%%%%%% Figure Start
\begin{figure}
\centering
{\graphicspath{{figs/figf/}}\includegraphics[width=7cm]{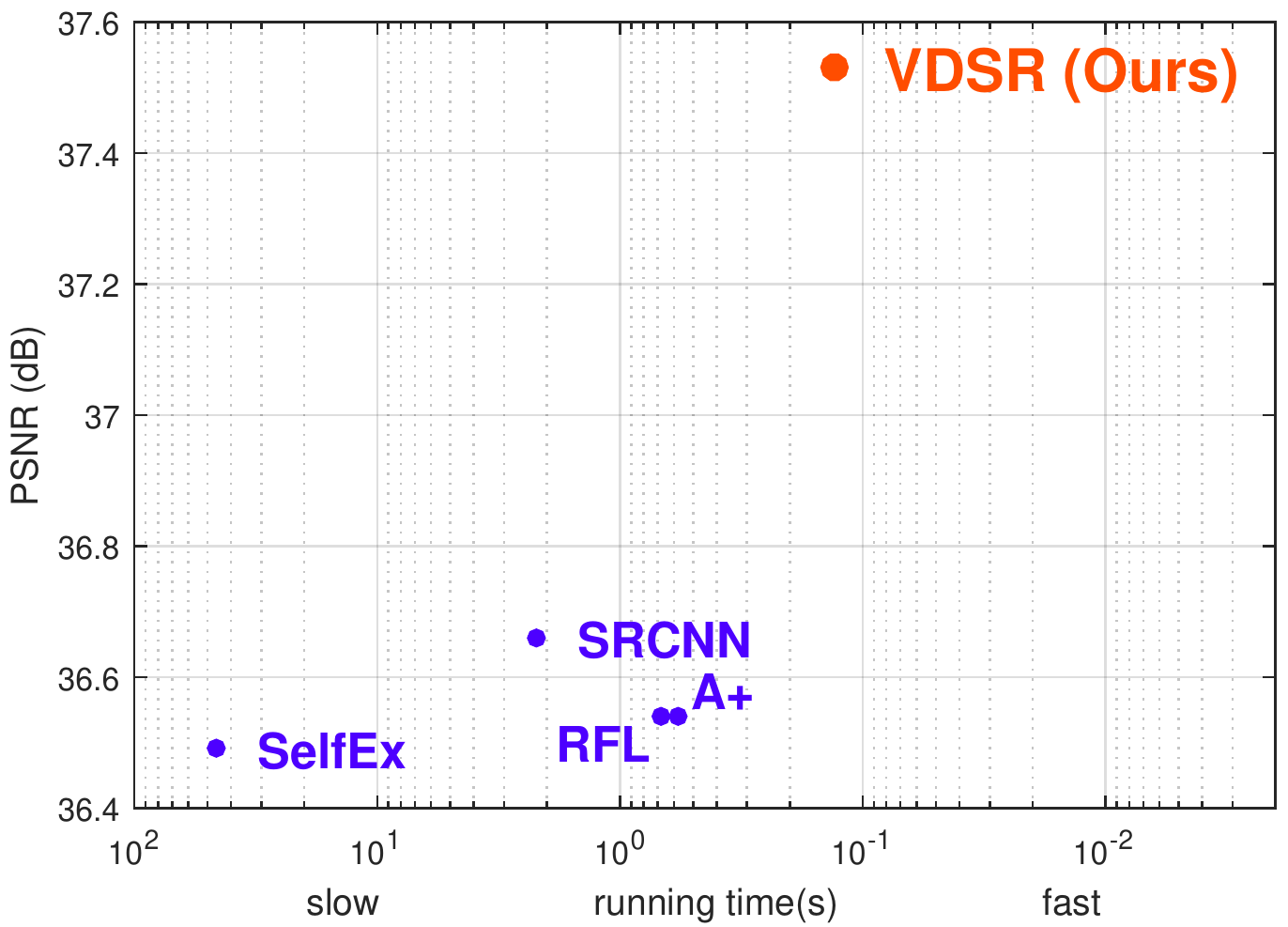}}
\caption{Our VDSR improves PSNR for scale factor $\times2$ on dataset Set5 in comparison to the state-of-the-art methods (SRCNN uses the public slower implementation using CPU). VDSR outperforms SRCNN by a large margin (0.87 dB).}
\label{fig:method_comparison}
\end{figure}
%%%%%%%%%% Figure End

In this work, we propose a new method to practically resolve the issues.

\textbf{Context} We utilize contextual information spread over very large image regions. For a large scale factor, it is often the case that information contained in a small patch is not sufficient for detail recovery (ill-posed). Our very deep network using large receptive field takes a large image context into account.

\textbf{Convergence} We suggest a way to speed-up the training: residual-learning CNN and extremely high learning rates. As LR image and HR image share the same information to a large extent, explicitly modelling the residual image, which is the difference between HR and LR images, is advantageous. We propose a network structure for efficient learning when input and output are highly correlated. Moreover, our initial learning rate is $10^4$ times higher than that of SRCNN \cite{dong2015image}. This is enabled by residual-learning and gradient clipping.

\textbf{Scale Factor} We propose a single-model SR approach. Scales are typically user-specified and can be arbitrary including fractions. For example, one might need smooth zoom-in in an image viewer or resizing to a specific dimension. Training and storing many scale-dependent models in preparation for all possible scenarios is impractical. We find a single convolutional network is sufficient for multi-scale-factor super-resolution.

\textbf{Contribution} In summary, in this work, we propose a highly accurate SR method based on a very deep convolutional network. Very deep networks converge too slowly if small learning rates are used. Boosting convergence rate with high learning rates lead to exploding gradients and we resolve the issue with residual-learning and gradient clipping. In addition, we extend our work to cope with multi-scale SR problem in a single network. Our method is relatively accurate and fast in comparison to state-of-the-art methods as illustrated in Figure \ref{fig:method_comparison}.

\section{Related Work}
SRCNN is a representative state-of-art method for deep learning-based SR approach. So, let us analyze and compare it with our proposed method.
\begin{figure*}[t]
\includegraphics[width=\textwidth]{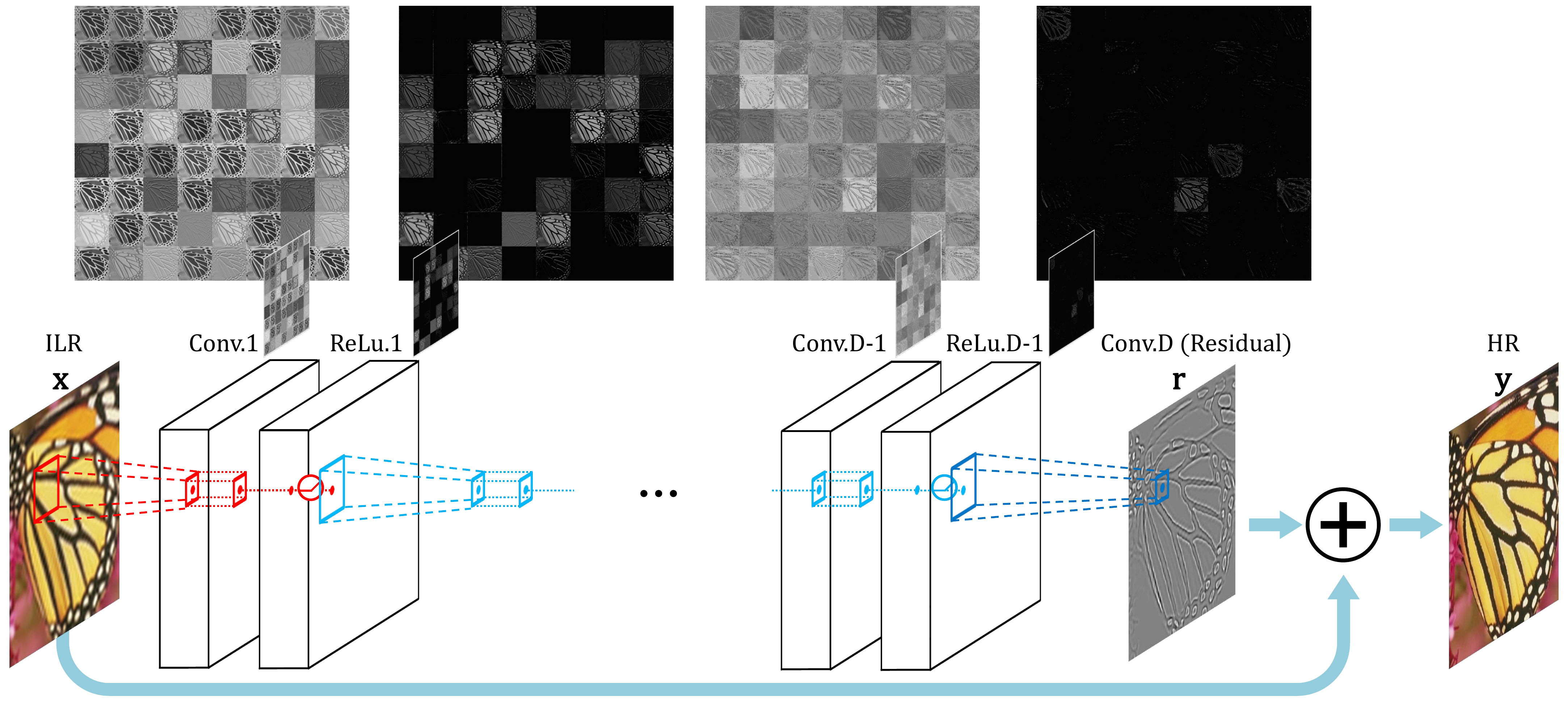}
\caption{Our Network Structure. We cascade a pair of layers (convolutional and nonlinear) repeatedly. An interpolated low-resolution (ILR) image goes through layers and transforms into a high-resolution (HR) image. The network predicts a residual image and the addition of ILR and the residual gives the desired output. We use 64 filters for each convolutional layer and some sample feature maps are drawn for visualization. Most features after applying rectified linear units (ReLu) are zero.}
\label{fig:network}
\end{figure*}

\subsection{Convolutional Network for Image Super-Resolution}
\textbf{Model}
SRCNN consists of three layers: patch extraction/representation, non-linear mapping and reconstruction. Filters of spatial sizes $9\times9$, $1\times1$, and $5\times5$ were used respectively. 

In \cite{dong2015image}, Dong et al. attempted to prepare deeper models, but failed to observe superior performance after a week of training. In some cases, deeper models gave inferior performance. They conclude that deeper networks do not result in better performance (Figure 9).

However, we argue that increasing depth significantly boosts performance. We successfully use 20 weight layers ($3\times3$ for each layer). Our network is very deep (20 vs. 3 \cite{dong2015image}) and information used for reconstruction (receptive field) is much larger ($41\times41$ vs. $13\times13$).

\textbf{Training}
For training, SRCNN directly models high-resolution images. A high-resolution image can be decomposed into a low frequency information (corresponding to low-resolution image) and high frequency information (residual image or image details). Input and output images share the same low-frequency information. This indicates that SRCNN  serves two purposes: carrying the input to the end layer and reconstructing residuals. Carrying the input to the end is conceptually similar to what an auto-encoder does. Training time might be spent on learning this auto-encoder so that the convergence rate of learning the other part (image details) is significantly decreased. In contrast, since our network models the residual images directly, we can have much faster convergence with even better accuracy.

\textbf{Scale} As in most existing SR methods, SRCNN is trained for a single scale factor and is supposed to work only with the specified scale. Thus, if a new scale is on demand, a new model has to be trained. To cope with multiple scale SR (possibly including fractional factors), we need to construct individual single scale SR system for each scale of interest.

However, preparing many individual machines for all possible scenarios to cope with multiple scales is inefficient and impractical.
In this work, we design and train a single network to handle multiple scale SR problem efficiently. This turns out to work very well. Our single machine is compared favorably to a single-scale expert for the given sub-task. For three scales factors ($\times 2,3,4$), we can reduce the number of parameters by three-fold.

In addition to the aforementioned issues, there are some minor differences. Our output image has the same size as the input image by padding zeros every layer during training whereas output from SRCNN is smaller than the input. Finally, we simply use the same learning rates for all layers while SRCNN uses different learning rates for different layers in order to achieve stable convergence.

\section{Proposed Method}
\subsection{Proposed Network}

For SR image reconstruction, we use a very deep convolutional network inspired by Simonyan and Zisserman \cite{simonyan2015very}. The configuration is outlined in Figure \ref{fig:network}. We use $d$ layers where layers except the first and the last are of the same type: 64 filter of the size $3\times 3 \times64$, where a filter operates on $3\times3$ spatial region across 64 channels (feature maps). The first layer operates on the input image. The last layer, used for image reconstruction, consists of a single filter of size $3\times 3 \times64$.

The network takes an interpolated low-resolution image (to the desired size) as input and predicts image details. Modelling image details is often used in super-resolution methods \cite{Timofte2013, Timofte, bevilacqua2012,bevilacqua2013super} and we find that CNN-based methods can benefit from this domain-specific knowledge.

In this work, we demonstrate that explicitly modelling image details (residuals) has several advantages. These are further discussed later in Section \ref{sec:residual}. 

One problem with using a very deep network to predict dense outputs is that the size of the feature map gets reduced every time convolution operations are applied. For example,  when an input of size $(n+1)\times (n+1)$ is applied to a network with receptive field size $n\times n$, the output image is $1\times1$. 

This is in accordance with other super-resolution methods since many require surrounding pixels to infer center pixels correctly. This center-surround relation is useful since the surrounding region provides more constraints to this ill-posed problem (SR). For pixels near the image boundary, this relation cannot be exploited to the full extent and many SR methods crop the result image. 

This methodology, however, is not valid if the required surround region is very big. After cropping, the final image is too small to be visually pleasing.

To resolve this issue, we pad zeros before convolutions to keep the sizes of all feature maps (including the output image) the same. It turns out that zero-padding works surprisingly well. For this reason, our method differs from most other methods in the sense that pixels near the image boundary are also correctly predicted.  

Once image details are predicted, they are added back to the input ILR image to give the final image (HR). We use this structure for all experiments in our work.

\subsection{Training}

We now describe the objective to minimize in order to find optimal parameters of our model. Let ${\bf x}$ denote an interpolated low-resolution image and ${\bf y}$ a high-resolution image. 
Given a training dataset $\{{\bf x}^{(i)},{\bf y}^{(i)}\}{}_{i=1}^{N}$, our goal is to learn a model $f$ that predicts values $\mathbf{\hat{y}}=f(\mathbf{x})$, where $\mathbf{\hat{y}}$ is an estimate of the target HR image. We minimize the mean squared error $\frac{1}{2}||\mathbf{y}-f(\mathbf{x})||^{2}$
averaged over the training set is minimized. 

\textbf{Residual-Learning}  In SRCNN, the exact copy of the input has to go through all layers until it reaches the output layer. With many weight layers, this becomes an end-to-end relation requiring very long-term memory. For this reason, the vanishing/exploding gradients problem \cite{bengio1994learning} can be critical. We can solve this problem simply with residual-learning.

As the input and output images are largely similar, we define a residual image ${\bf r}={\bf y}-{\bf x}$, where most values are likely to be zero or small. We want to predict this residual image. The loss function now becomes $\frac{1}{2}||\mathbf{r}-f(\mathbf{x})||^{2}$, where $f(\bf{x})$ is the network prediction. 

In networks, this is reflected in the loss layer as follows. 
Our loss layer takes three inputs: residual estimate, network input (ILR image) and ground truth HR image. The loss is computed as the Euclidean distance between the reconstructed image (the sum of network input and output) and ground truth. 

Training is carried out by optimizing the regression objective using mini-batch gradient descent based on back-propagation (LeCun et al. \cite{lecun1998gradient}). We set the momentum parameter to 0.9. The training is regularized by weight decay ($L_2$ penalty multiplied by
0.0001).  

\textbf{High Learning Rates for Very Deep Networks}
Training deep models can fail to converge in realistic limit of time. SRCNN \cite{dong2015image} fails to show superior performance with more than three weight layers. While there can be various reasons, one possibility is that they stopped their training procedure before networks converged. Their learning rate $10^{-5}$ is too small for a network to converge within a week on a common GPU. Looking at Fig. 9 of \cite{dong2015image}, it is not easy to say their deeper networks have converged and their performances were saturated. While more training will eventually resolve the issue, but increasing depth to 20 does not seems practical with SRCNN.

It is a basic rule of thumb to make learning rate high to boost training. But simply setting learning rate high can also lead to vanishing/exploding gradients \cite{bengio1994learning}. For the reason, we suggest an adjustable gradient clipping for maximal boost in speed while suppressing exploding gradients.

\textbf{Adjustable Gradient Clipping}
Gradient clipping is a technique that is often used in training recurrent neural networks \cite{pascanu2013difficulty}. But, to our knowledge, its usage is limited in training CNNs. While there exist many ways to limit gradients, one of the common strategies is to clip individual gradients to the predefined range 
$[-\theta, \theta]$. 

With clipping, gradients are in a certain range. With stochastic gradient descent commonly used for training, learning rate is multiplied to adjust the step size. If high learning rate is used, it is likely that $\theta$ is tuned to be small to avoid exploding gradients in a high learning rate regime. But as learning rate is annealed to get smaller, the effective gradient (gradient multiplied by learning rate) approaches zero and training can take exponentially many iterations to converge if learning rate is decreased geometrically.

For maximal speed of convergence, we clip the gradients to $[-\frac{\theta}{\gamma}, \frac{\theta}{\gamma}]$, where $\gamma$ denotes the current learning rate. We find the adjustable gradient clipping makes our convergence procedure extremely fast. Our 20-layer network training is done within 4 hours whereas 3-layer SRCNN takes several days to train.

\textbf{Multi-Scale} While very deep models can boost performance, more parameters are now needed to define a network. Typically, one network is created for each scale factor. Considering that fractional scale factors are often used, we need an economical way to store and retrieve networks.

For this reason, we also train a multi-scale model. With this approach, parameters are shared across all predefined scale factors. Training a multi-scale model is straightforward. Training datasets for several specified scales are combined into one big dataset.

Data preparation is similar to SRCNN \cite{Dong2014} with some differences. Input patch size is now equal to the size of the receptive field and images are divided into sub-images with no overlap. A mini-batch consists of 64 sub-images, where sub-images from different scales can be in the same batch.

We implement our model using the \textit{MatConvNet}\footnote{\url{ http://www.vlfeat.org/matconvnet/}} package \cite{arXiv:1412.4564}.

\begin{figure*}[t]
\vspace{-.5cm}
	\centering
	\begin{subfigure}{0.25\textwidth}
		\includegraphics[width=\textwidth]{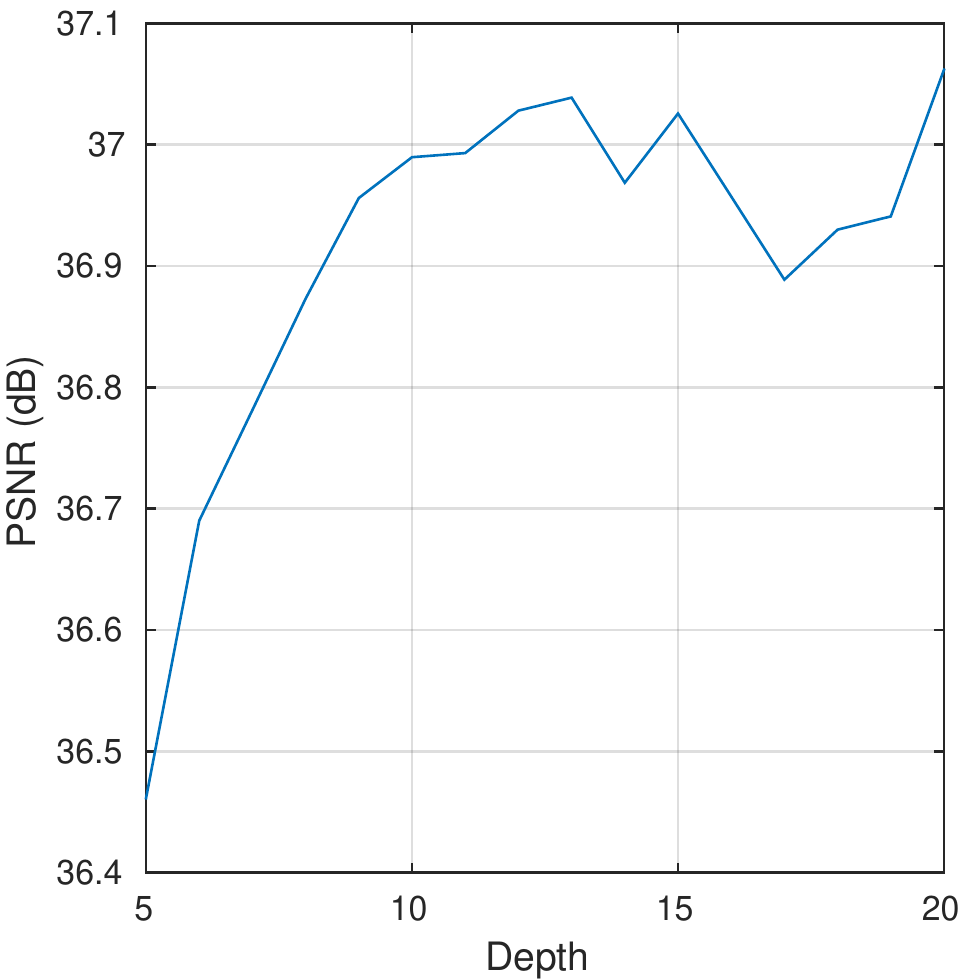}
		\caption{Test Scale Factor 2}
		\label{fig:gull}
	\end{subfigure}%
	\quad
	\begin{subfigure}{0.25\textwidth}
		\includegraphics[width=\textwidth]{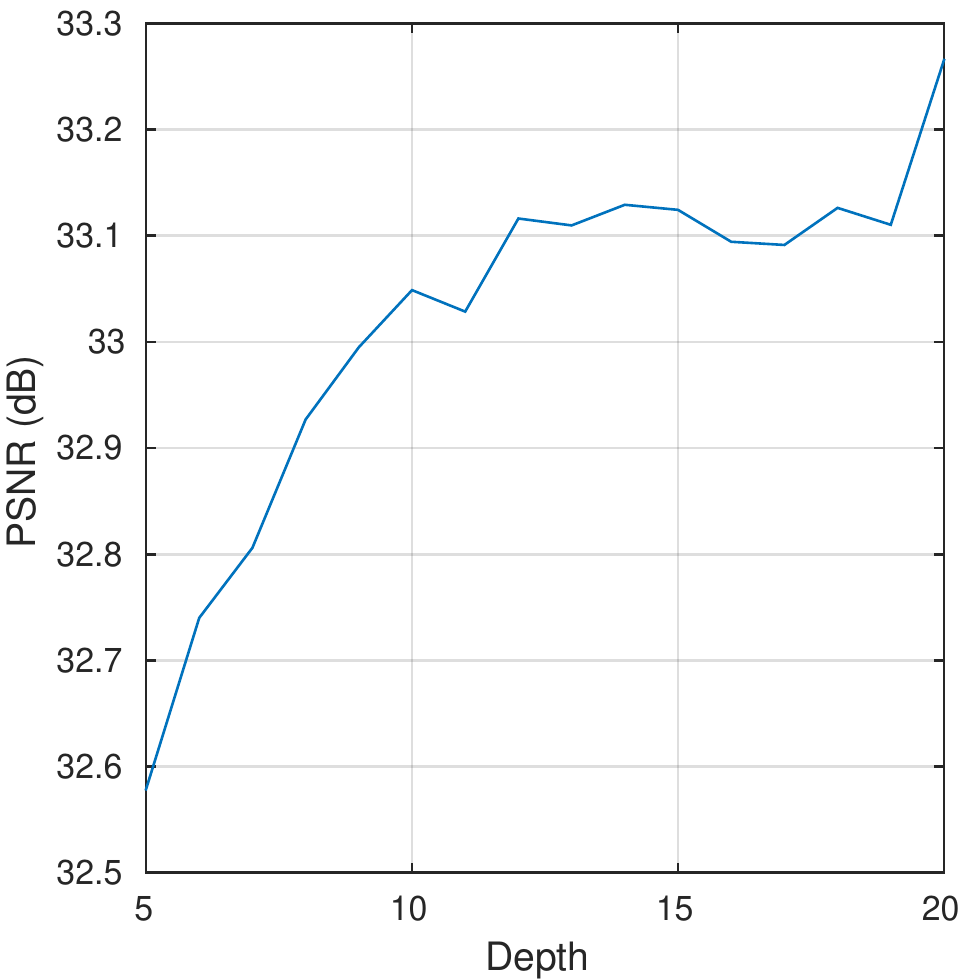}
		\caption{Test Scale Factor 3}
		\label{fig:tiger}
	\end{subfigure}
	\quad
	\begin{subfigure}{0.25\textwidth}
		\includegraphics[width=\textwidth]{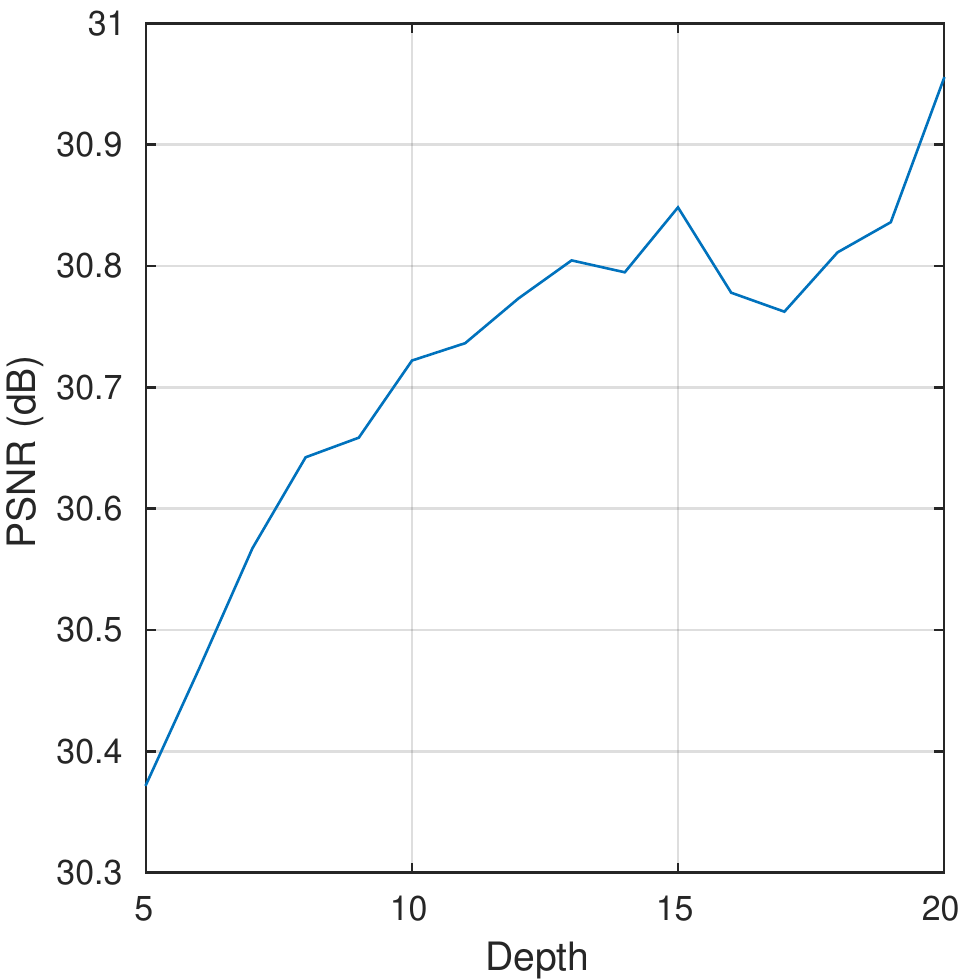}
		\caption{Test Scale Factor 4}
		\label{fig:mouse}
	\end{subfigure}
	\caption{Depth vs Performance}\label{fig:depth}
\end{figure*}

\begin{table}
\vspace{-.5cm}
\centering
\begin{subtable}[t]{0.9\linewidth}
\centering
\begin{tabular}{|c||c|c|c|c|}
\hline
 Epoch& 10& 20 & 40 & 80\\\hline
 Residual& \color{red}36.90& 36.64& 37.12&  37.05\\
 Non-Residual&  \color{red}27.42& 19.59& 31.38& 35.66\\\hline
 Difference&  \color{red} 9.48& 17.05& 5.74 & 1.39\\\hline
\end{tabular}
\caption{Initial learning rate 0.1}
\end{subtable}
\begin{subtable}[t]{0.9\linewidth}
\centering
\begin{tabular}{|c||c|c|c|c|}
\hline
 Epoch& 10& 20 & 40 & 80\\\hline
 Residual& \color{red}36.74& 36.87& 36.91&  36.93\\
 Non-Residual& \color{red}30.33& 33.59& 36.26&  36.42\\\hline
 Difference& \color{red}6.41& 3.28& 0.65& 0.52\\\hline
\end{tabular}
\caption{Initial learning rate 0.01}
\end{subtable}
\begin{subtable}[t]{0.9\linewidth}
\centering
\begin{tabular}{|c||c|c|c|c|}
\hline
 Epoch& 10& 20 & 40 & 80\\\hline
 Residual& \color{red}36.31& 36.46& 36.52& 36.52\\
 Non-Residual& \color{red}33.97& 35.08& 36.11&  36.11\\\hline
 Difference& \color{red}2.35& 1.38& 0.42& 0.40\\\hline
\end{tabular}
\caption{Initial learning rate 0.001}
\end{subtable}
\caption{Performance table (PSNR) for residual and non-residual networks (`Set5' dataset, $\times$ 2). Residual networks rapidly approach their convergence within 10 epochs.}
\end{table}

\begin{figure*}[t]
\vspace{-.3cm}
	\centering
	\begin{subfigure}{0.3\textwidth}
		\includegraphics[width=\textwidth]{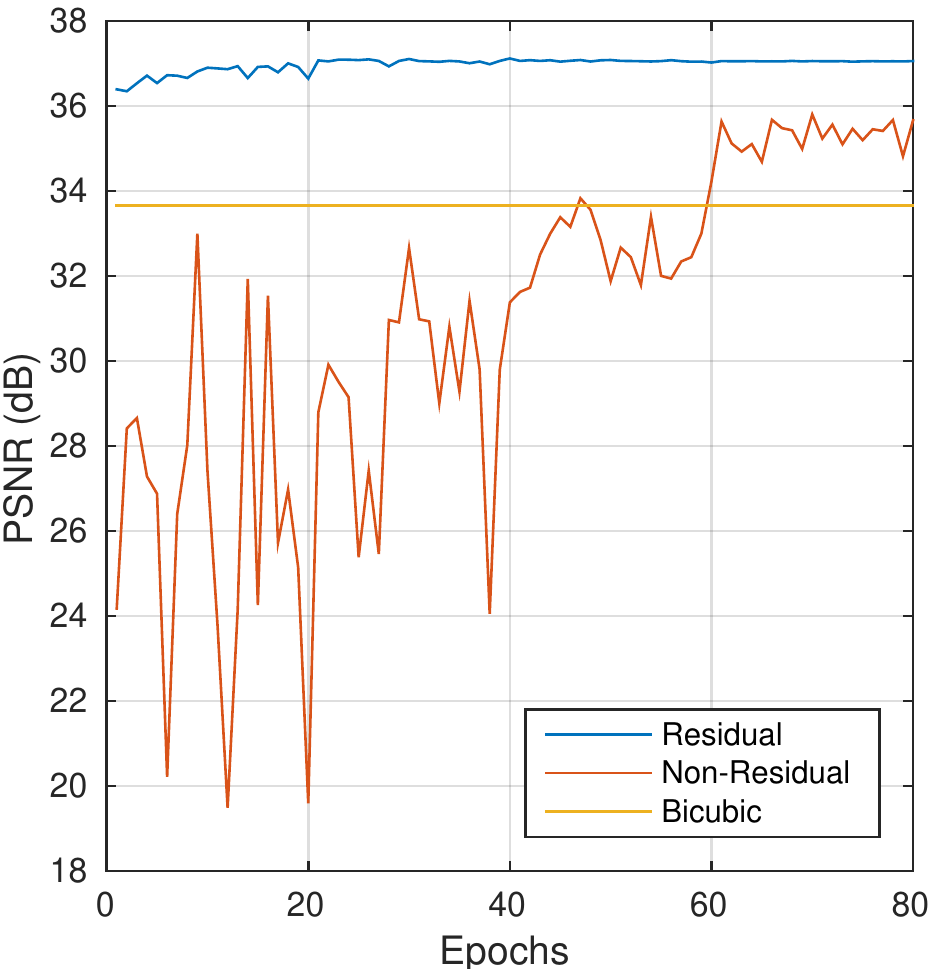}
		\caption{Initial learning rate 0.1}
		\label{fig:gull}
	\end{subfigure}%
	\hfill
	\begin{subfigure}{0.3\textwidth}
		\includegraphics[width=\textwidth]{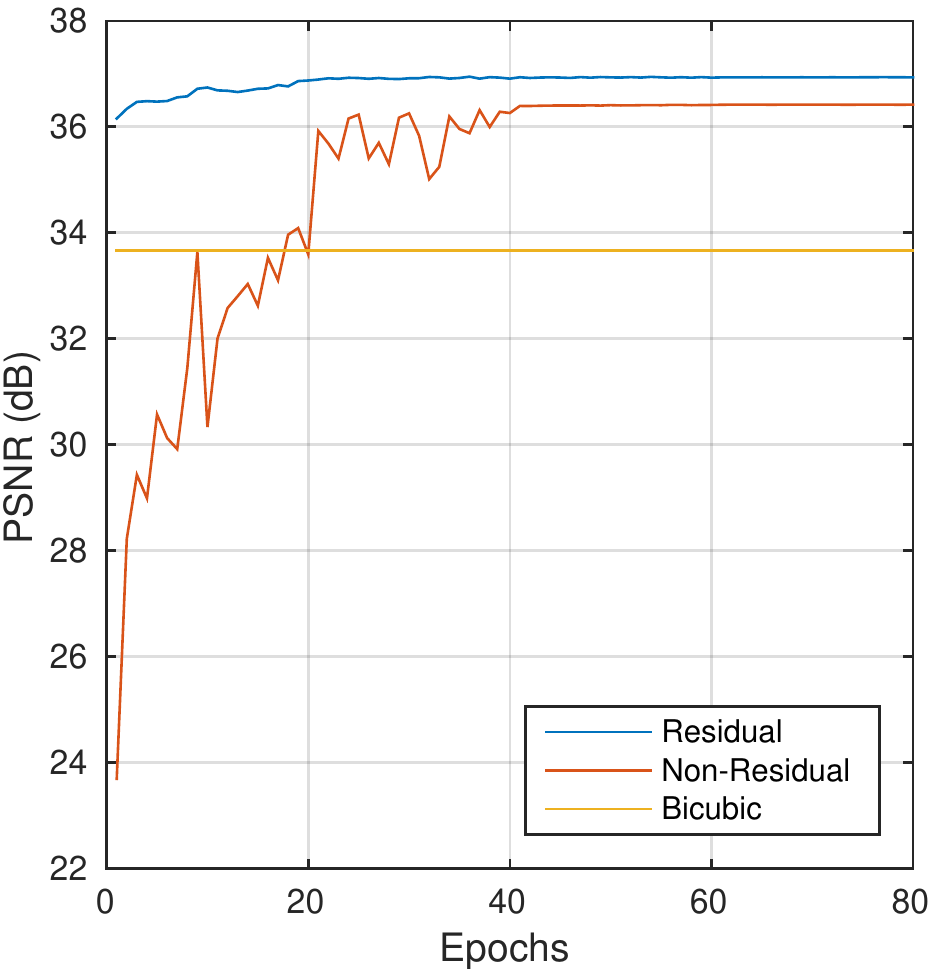}
		\caption{Initial learning rate 0.01}
		\label{fig:tiger}
	\end{subfigure}
	\hfill
	\begin{subfigure}{0.3\textwidth}
		\includegraphics[width=\textwidth]{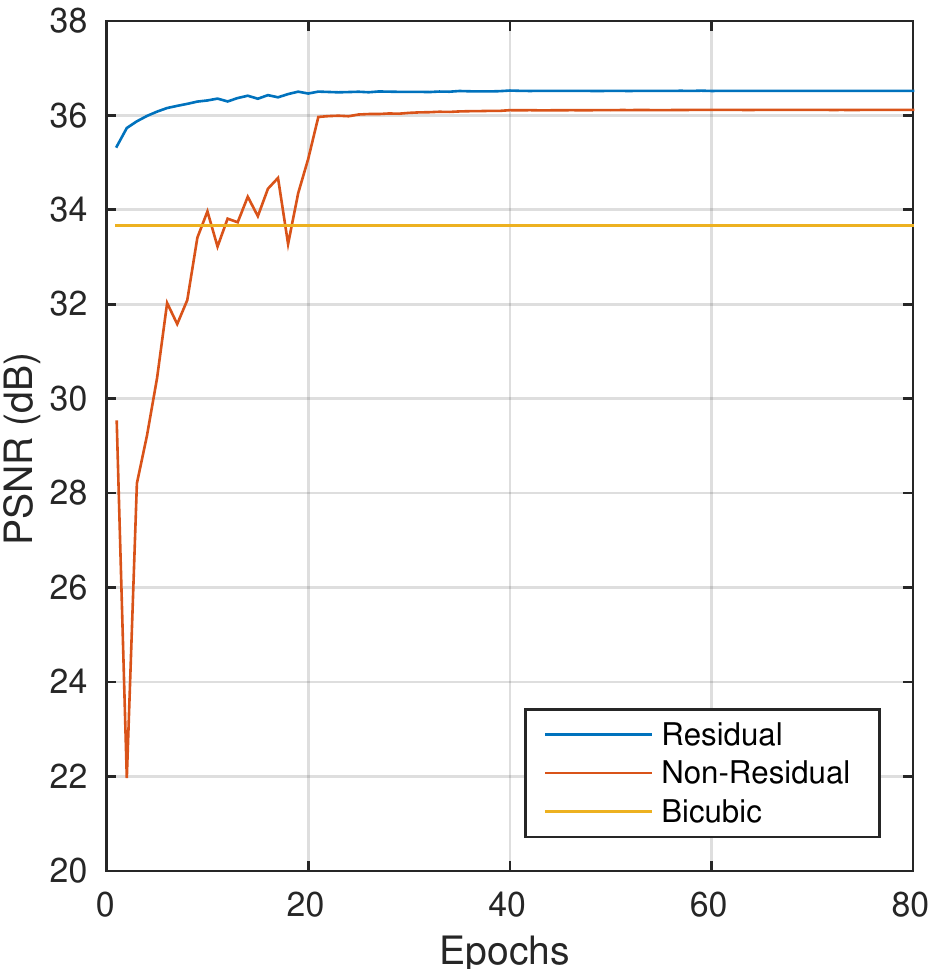}
		\caption{Initial learning rate 0.001}
		\label{fig:mouse}
	\end{subfigure}
	\caption{Performance curve for residual and non-residual networks. Two networks are tested under `Set5' dataset with scale factor 2. Residual networks quickly reach state-of-the-art performance within a few epochs, whereas non-residual networks (which models high-resolution image directly) take many epochs to reach maximum performance. Moreover, the final accuracy is higher for residual networks.}
	\label{fig:residual2}
\end{figure*}

\begin{table*}[t]
	\small
	\centering
\begin{tabular}
{|c|c|c|c|c|c|c|c||c|}
\hline 
 Test / Train & {$\times$2}& {$\times$3}& { $\times$4}& {$\times$2,3}& {$\times$2,4}& { $\times$3,4}& {$\times$2,3,4} & {Bicubic} \\
\hline
$\times$2  & \color{red} 37.10  & 30.05  & 28.13  & \color{red} 37.09  & \color{red} 37.03  & 32.43  & \color{red}37.06 &33.66   \\
$\times$3  & 30.42  & \color{red} 32.89  & 30.50  & \color{red} 33.22  & 31.20  & \color{red} 33.24  & \color{red} 33.27  & 30.39 \\
$\times$4  & 28.43  & 28.73  & \color{red} 30.84  & 28.70  & \color{red} 30.86  & \color{red} 30.94  & \color{red} 30.95 & 28.42  \\
\hline
\end{tabular}
	\vspace{1pt}
	\caption{Scale Factor Experiment. Several models are trained with different scale sets. Quantitative evaluation (PSNR) on dataset `Set5' is provided for scale factors 2,3 and 4.  {\color{red}Red color} indicates that test scale is included during training. Models trained with multiple scales perform well on the trained scales. }
	\label{tab:SRCNN_Factor_Test}
\end{table*}

\begin{figure*}
\vspace{-.5cm}
\includegraphics[width=\textwidth]{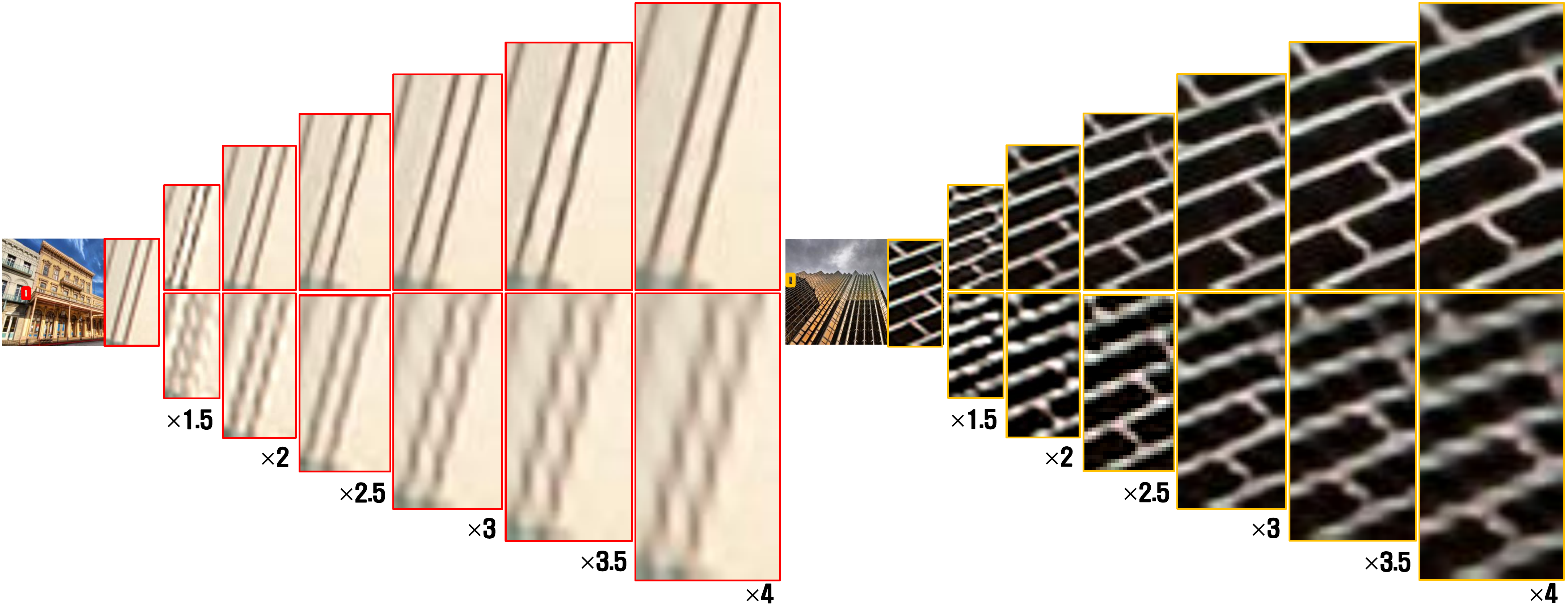}
\caption{(Top) Our results using a single network for all scale factors. Super-resolved images over all scales are clean and sharp. (Bottom)  Results of Dong et al.  \cite{Dong2014} ($\times$3 model used for all scales). Result images are not visually pleasing. To handle multiple scales, existing methods require multiple networks.}
\end{figure*}

\section{Understanding Properties}

In this section, we study three properties of our proposed method. First, we show that large depth is necessary for the task of SR. A very deep network utilizes more contextual information in an image and models complex functions with many nonlinear layers. We experimentally verify that deeper networks give better performances than shallow ones. 

Second, we show that our residual-learning network converges much faster than the standard CNN. Moreover, our network gives a significant boost in performance. 

Third, we show that our method with a single network performs as well as a method using multiple networks trained for each scale. We can effectively reduce model capacity (the number of parameters) of multi-network approaches.

\subsection{The Deeper, the Better}
Convolutional neural networks exploit spatially-local correlation by enforcing a local connectivity pattern between neurons of adjacent layers \cite{Bengio-et-al-2015-Book}. In other words, hidden units in layer $m$ take as input a subset of units in layer $m-1$. They form spatially contiguous receptive fields.

Each hidden unit is unresponsive to variations outside of the receptive field with respect to the input. The architecture thus ensures that the learned filters produce the strongest response to a spatially local input pattern.

However, stacking many such layers leads to filters that become increasingly “global” (i.e. responsive to a larger region of pixel space). In other words, a filter of very large support can be effectively decomposed into a series of small filters. 

In this work, we use filters of the same size, 3$\times$3, for all layers. For the first layer, the receptive field is of size 3$\times$3. For the next layers, the size of the receptive field increases by 2 in both height and width. For depth $D$ network, the receptive field has size $(2D+1)\times(2D+1)$. Its size is proportional to the depth.

In the task of SR, this corresponds to the amount of contextual information that can be exploited to infer high-frequency components. A large receptive field means the network can use more context to predict image details. As SR is an ill-posed inverse problem, collecting and analyzing more neighbor pixels give more clues. For example, if there are some image patterns entirely contained in a receptive field, it is plausible that this pattern is recognized and used to super-resolve the image. 

In addition, very deep networks can exploit high nonlinearities. We use 19 rectified linear units and our networks can model very complex functions with moderate number of channels (neurons). The advantages of making a thin deep network is well explained in Simonyan and Zisserman \cite{simonyan2015very}.

We now experimentally show that very deep networks significantly improve SR performance. We train and test networks of depth ranging from 5 to 20 (only counting weight layers excluding nonlinearity layers). In Figure \ref{fig:depth}, we show the results. In most cases, performance increases as depth increases. As depth increases, performance improves rapidly. 
%
%\begin{figure*}
%\vspace{-.5cm}
%\begin{adjustwidth}{0cm}{-0.5cm}
%\begin{center}
%\footnotesize
%\setlength{\tabcolsep}{5pt}
%\begin{tabular}{ c C{3.5cm}  C{3.5cm}  C{3.5cm}  }
%\multirow{4}{*}{\graphicspath{{figs/fig1/}}\includegraphics[width=0.17\textwidth]{img039_GTbox.png}}
%& \raisebox{-13.0ex} {\graphicspath{{figs/fig1/}}\includegraphics[width=0.20\textwidth]{img039_for_fig1_HR.png}}\vspace{0.3ex}
%& \raisebox{-13.0ex} {\graphicspath{{figs/fig1/}}\includegraphics[width=0.20\textwidth]{img039_for_fig1_A+.png}}\vspace{0.3ex}
%& \raisebox{-13.0ex} {\graphicspath{{figs/fig1/}}\includegraphics[width=0.20\textwidth]{img039_for_fig1_RFL.png}}\vspace{0.3ex}
%\\
%& Original (PSNR, SSIM)& A+ (22.74, 0.7234)& RFL (22.54, 0.7092)\\
%& \raisebox{-13.0ex} {\graphicspath{{figs/fig1/}}\includegraphics[width=0.20\textwidth]{img039_for_fig1_SelfEx.png}}\vspace{0.3ex}
%& \raisebox{-13.0ex} {\graphicspath{{figs/fig1/}}\includegraphics[width=0.20\textwidth]{img039_for_fig1_SRCNN.png}}\vspace{0.3ex}
%& \raisebox{-13.0ex} {\graphicspath{{figs/fig1/}}\includegraphics[width=0.20\textwidth]{img039_for_fig1_VDSR.png}}\vspace{0.3ex}
%\\
%& SelfEx (22.79, {\color{blue}{0.7341}})& SRCNN ({\color{blue}{22.84}}, 0.7248)& VDSR (Ours) ({\color{red}{23.91}}, {\color{red}{0.7858}})\\
%\end{tabular}
%\caption{Super-resolution results of ``img039"(Urban100) with scale factor $\times$3. Our result is visually pleasing.}
%\label{fig:c1}
%\end{center}
%\end{adjustwidth}
%\end{figure*}

\begin{figure*}
\begin{adjustwidth}{0.5cm}{0.5cm}
\begin{center}
\small
\setlength{\tabcolsep}{3pt}
\begin{tabular}{  c  c  c  c  c  c  }
{\graphicspath{{figs/figVDSR/}}\includegraphics[width=0.15\textwidth]{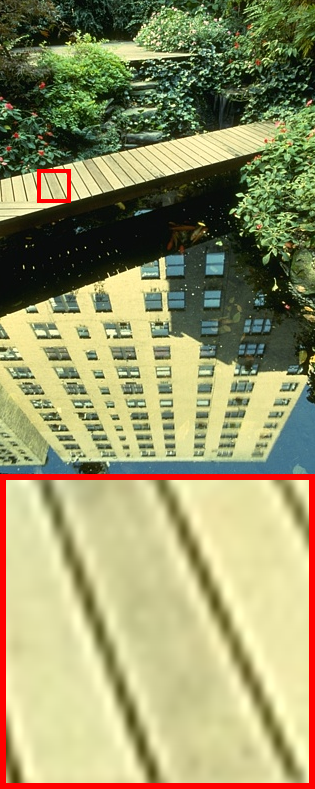}}
& {\graphicspath{{figs/figVDSR/}}\includegraphics[width=0.15\textwidth]{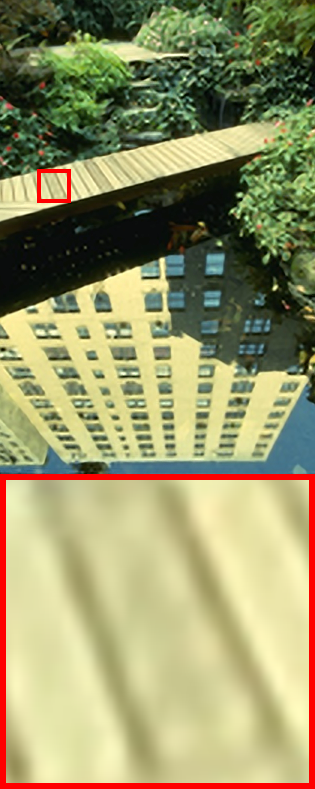}}
& {\graphicspath{{figs/figVDSR/}}\includegraphics[width=0.15\textwidth]{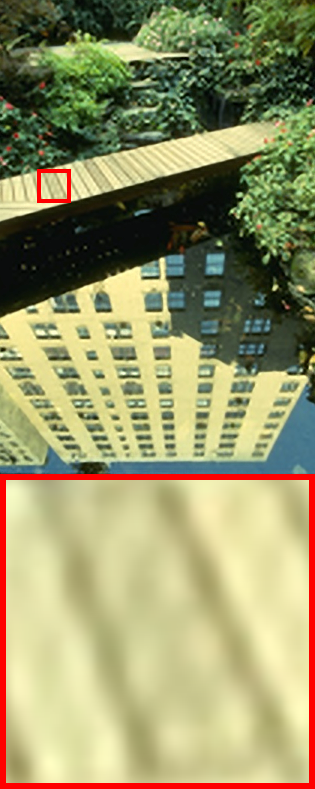}}
& {\graphicspath{{figs/figVDSR/}}\includegraphics[width=0.15\textwidth]{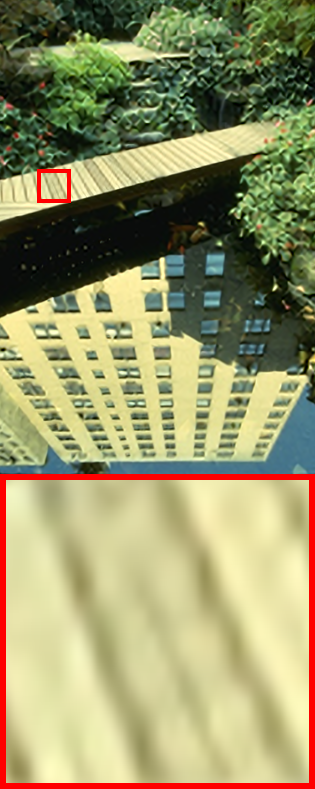}}
& {\graphicspath{{figs/figVDSR/}}\includegraphics[width=0.15\textwidth]{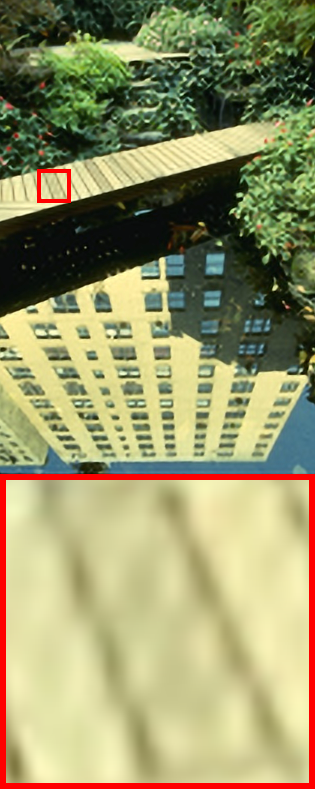}}
& {\graphicspath{{figs/figVDSR/}}\includegraphics[width=0.15\textwidth]{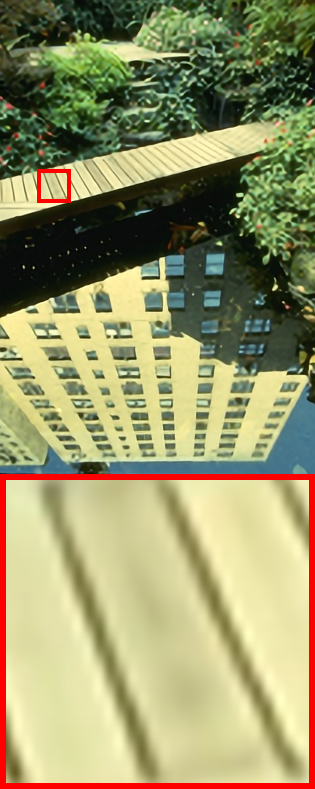}}
\\
Ground Truth& A+ \cite{Timofte}& RFL \cite{schulter2015fast}& SelfEx \cite{Huang-CVPR-2015}& SRCNN \cite{Dong2014}& VDSR (Ours) \\
(PSNR, SSIM)& (22.92, 0.7379)& (22.90, 0.7332)& (23.00, 0.7439)& ({\color{blue}{23.15}}, {\color{blue}{0.7487}})& ({\color{red}{23.50}}, {\color{red}{0.7777}})\\
\end{tabular}
\caption{Super-resolution results of ``148026" (\textit{B100}) with scale factor $\times$3. VDSR recovers sharp lines. }
\label{fig:c2}
\end{center}
\end{adjustwidth}
\end{figure*}

\begin{figure*}
\begin{adjustwidth}{0.5cm}{0.5cm}
\begin{center}
\small
\setlength{\tabcolsep}{3pt}
\begin{tabular}{  c  c  c  c  c  c  }
{\graphicspath{{figs/figVDSR/}}\includegraphics[width=0.15\textwidth]{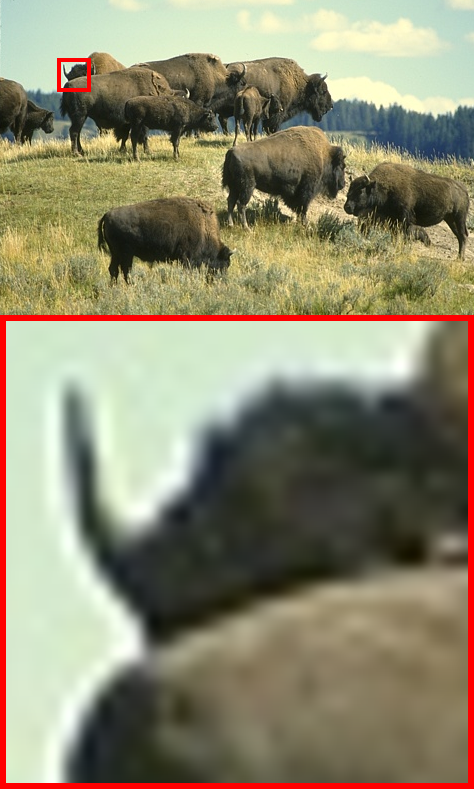}}
& {\graphicspath{{figs/figVDSR/}}\includegraphics[width=0.15\textwidth]{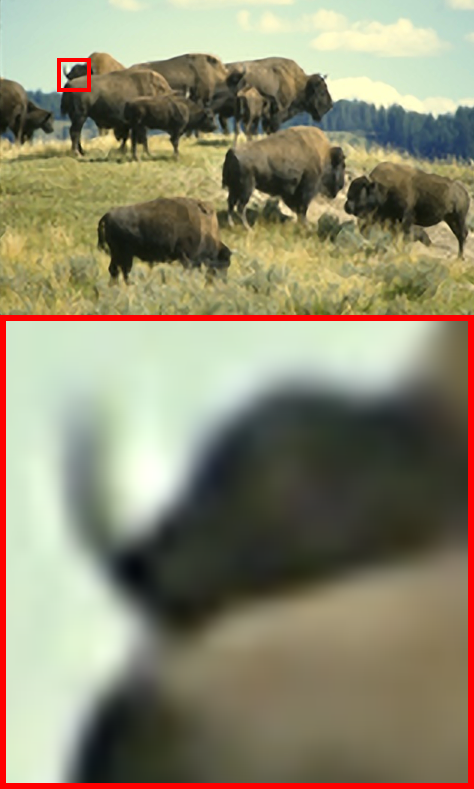}}
& {\graphicspath{{figs/figVDSR/}}\includegraphics[width=0.15\textwidth]{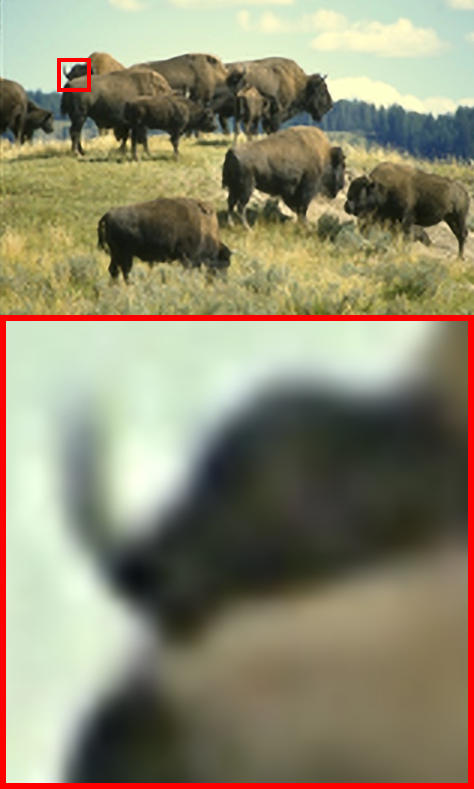}}
& {\graphicspath{{figs/figVDSR/}}\includegraphics[width=0.15\textwidth]{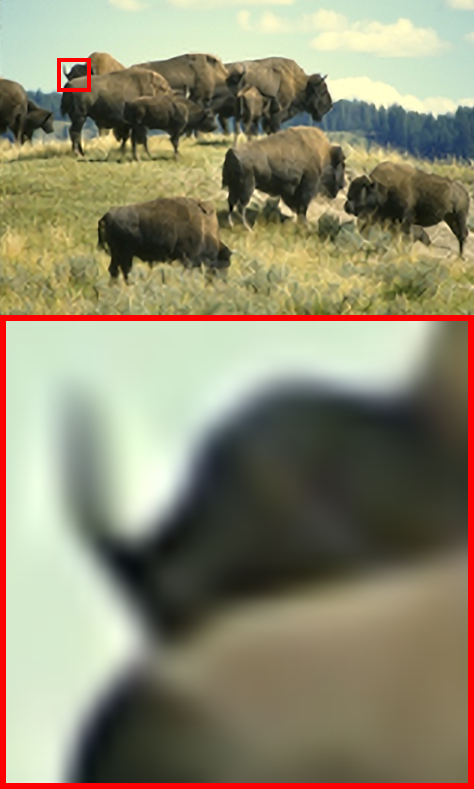}}
& {\graphicspath{{figs/figVDSR/}}\includegraphics[width=0.15\textwidth]{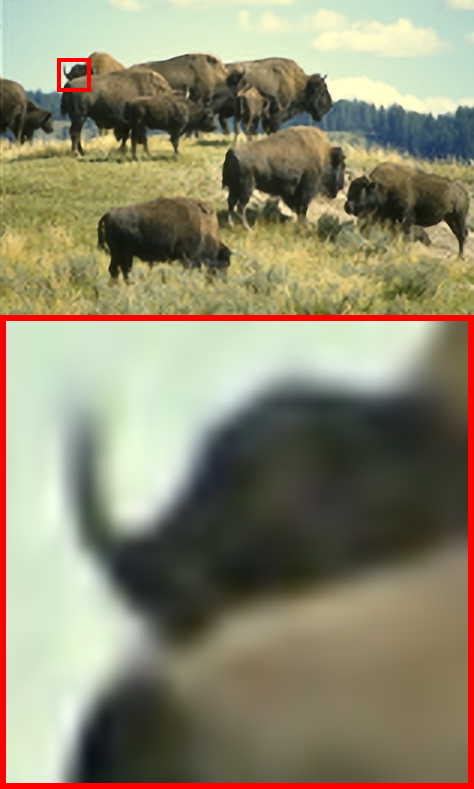}}
& {\graphicspath{{figs/figVDSR/}}\includegraphics[width=0.15\textwidth]{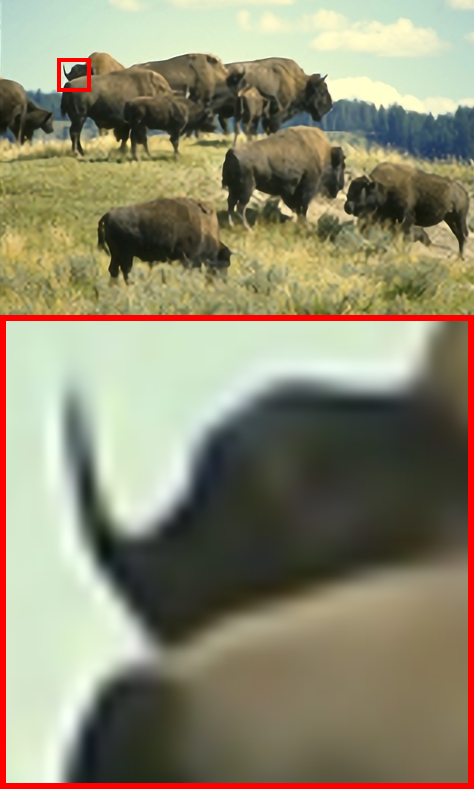}}
\\n 
Ground Truth& A+ \cite{Timofte}& RFL \cite{schulter2015fast}& SelfEx \cite{Huang-CVPR-2015}& SRCNN \cite{Dong2014}& VDSR (Ours) \\
(PSNR, SSIM)& (27.08, 0.7514)& (27.08, 0.7508)& (27.02, 0.7513)& ({\color{blue}{27.16}}, {\color{blue}{0.7545}})& ({\color{red}{27.32}}, {\color{red}{0.7606}})\\
\end{tabular}
\caption{Super-resolution results of ``38092" (\textit{B100}) with scale factor $\times$3. The horn in the image is sharp in the result of VDSR.}
\label{fig:c3}
\end{center}
\end{adjustwidth}
\end{figure*}

\begin{table*}
\begin{center}
\setlength{\tabcolsep}{2pt}
\small
\begin{tabular}{ | c | c | c | c | c | c | c | c | }
\hline
\multirow{2}{*}{Dataset} & \multirow{2}{*}{Scale} & Bicubic & A+ \cite{Timofte} & RFL \cite{schulter2015fast} & SelfEx \cite{Huang-CVPR-2015} & SRCNN \cite{Dong2014} & VDSR (Ours)\\
 & & PSNR/SSIM/time & PSNR/SSIM/time & PSNR/SSIM/time & PSNR/SSIM/time & PSNR/SSIM/time & PSNR/SSIM/time\\
\hline
\hline
\multirow{3}{*}{Set5} & $\times$2 & 33.66/0.9299/0.00 & 36.54/{\color{blue}0.9544}/{\color{blue}0.58} & 36.54/0.9537/0.63 & 36.49/0.9537/45.78 & {\color{blue}36.66}/0.9542/2.19 & {\color{red}37.53}/{\color{red}0.9587}/{\color{red}0.13}\\
 & $\times$3 & 30.39/0.8682/0.00 & 32.58/0.9088/{\color{blue}0.32} & 32.43/0.9057/0.49 & 32.58/{\color{blue}0.9093}/33.44 & {\color{blue}32.75}/0.9090/2.23 & {\color{red}33.66}/{\color{red}0.9213}/{\color{red}0.13}\\
 & $\times$4 & 28.42/0.8104/0.00 & 30.28/0.8603/{\color{blue}0.24} & 30.14/0.8548/0.38 & 30.31/0.8619/29.18 & {\color{blue}30.48}/{\color{blue}0.8628}/2.19 & {\color{red}31.35}/{\color{red}0.8838}/{\color{red}0.12}\\
\hline
\hline
\multirow{3}{*}{Set14} & $\times$2 & 30.24/0.8688/0.00 & 32.28/0.9056/{\color{blue}0.86} & 32.26/0.9040/1.13 & 32.22/0.9034/105.00 & {\color{blue}32.42}/{\color{blue}0.9063}/4.32 & {\color{red}33.03}/{\color{red}0.9124}/{\color{red}0.25}\\
 & $\times$3 & 27.55/0.7742/0.00 & 29.13/0.8188/{\color{blue}0.56} & 29.05/0.8164/0.85 & 29.16/0.8196/74.69 & {\color{blue}29.28}/{\color{blue}0.8209}/4.40 & {\color{red}29.77}/{\color{red}0.8314}/{\color{red}0.26}\\
 & $\times$4 & 26.00/0.7027/0.00 & 27.32/0.7491/{\color{blue}0.38} & 27.24/0.7451/0.65 & 27.40/{\color{blue}0.7518}/65.08 & {\color{blue}27.49}/0.7503/4.39 & {\color{red}28.01}/{\color{red}0.7674}/{\color{red}0.25}\\
\hline
\hline
\multirow{3}{*}{B100} & $\times$2 & 29.56/0.8431/0.00 & 31.21/0.8863/{\color{blue}0.59} & 31.16/0.8840/0.80 & 31.18/0.8855/60.09 & {\color{blue}31.36}/{\color{blue}0.8879}/2.51 & {\color{red}31.90}/{\color{red}0.8960}/{\color{red}0.16}\\
 & $\times$3 & 27.21/0.7385/0.00 & 28.29/0.7835/{\color{blue}0.33} & 28.22/0.7806/0.62 & 28.29/0.7840/40.01 & {\color{blue}28.41}/{\color{blue}0.7863}/2.58 & {\color{red}28.82}/{\color{red}0.7976}/{\color{red}0.21}\\
 & $\times$4 & 25.96/0.6675/0.00 & 26.82/0.7087/{\color{blue}0.26} & 26.75/0.7054/0.48 & 26.84/{\color{blue}0.7106}/35.87 & {\color{blue}26.90}/0.7101/2.51 & {\color{red}27.29}/{\color{red}0.7251}/{\color{red}0.21}\\
\hline
\hline
\multirow{3}{*}{Urban100} & $\times$2 & 26.88/0.8403/0.00 & 29.20/0.8938/{\color{blue}2.96} & 29.11/0.8904/3.62 & {\color{blue}29.54}/{\color{blue}0.8967}/663.98 & 29.50/0.8946/22.12 & {\color{red}30.76}/{\color{red}0.9140}/{\color{red}0.98}\\
 & $\times$3 & 24.46/0.7349/0.00 & 26.03/0.7973/{\color{blue}1.67} & 25.86/0.7900/2.48 & {\color{blue}26.44}/{\color{blue}0.8088}/473.60 & 26.24/0.7989/19.35 & {\color{red}27.14}/{\color{red}0.8279}/{\color{red}1.08}\\
 & $\times$4 & 23.14/0.6577/0.00 & 24.32/0.7183/{\color{blue}1.21} & 24.19/0.7096/1.88 & {\color{blue}24.79}/{\color{blue}0.7374}/394.40 & 24.52/0.7221/18.46 & {\color{red}25.18}/{\color{red}0.7524}/{\color{red}1.06}\\
\hline
\end{tabular}
\caption{Average PSNR/SSIM for scale factor $\times$2, $\times$3 and $\times$4 on datasets Set5, Set14, B100 and Urban100. {\color{red}Red color} indicates the best performance and {\color{blue}blue color} indicates the second best performance.}
\label{tbl:benchmark}
\end{center}
\end{table*}

\subsection{Residual-Learning}
\label{sec:residual}

As we already have a low-resolution image as the input, predicting high-frequency components is enough for the purpose of SR. Although the concept of predicting residuals has been used in previous methods \cite{Timofte2013, Timofte,zeyde2012single}, it has not been studied in the context of deep-learning-based SR framework.

In this work, we have proposed a network structure that learns residual images. We now study the effect of this modification to a standard CNN structure in detail. 

First, we find that this residual network converges much faster. Two networks are compared experimentally: the residual network and the standard non-residual network. We use depth 10 (weight layers) and scale factor 2. Performance curves for various learning rates are shown in Figure \ref{fig:residual2}. All use the same learning rate scheduling mechanism that has been mentioned above. 

Second, at convergence, the residual network shows superior performance. In Figure \ref{fig:residual2}, residual networks give higher PSNR when training is done.

Another remark is that if small learning rates are used, networks do not converge in the given number of epochs. If initial learning rate 0.1 is used, PSNR of a residual-learning network reaches 36.90 within 10 epochs. But if 0.001 is used instead, the network never reaches the same level of performance (its performance is 36.52 after 80 epochs). In a similar manner, residual and non-residual networks show dramatic performance gaps after 10 epochs (36.90 vs. 27.42 for rate 0.1).

In short, this simple modification to a standard non-residual network structure is very powerful and one can explore the validity of the idea in other image restoration problems where input and output images are highly correlated.

\subsection{Single Model for Multiple Scales}
Scale augmentation during training is a key technique to equip a network with super-resolution machines of multiple scales. Many SR processes for different scales can be executed with our multi-scale machine with much smaller capacity than that of single-scale machines combined. 

We start with an interesting experiment as follows: we train our network with a single scale factor $s_{\text{train}}$ and it is tested under another scale factor $s_{\text{test}}$. Here, factors 2,3 and 4 that are widely used in SR comparisons are considered. Possible pairs ($s_{\text{train}}$,$s_{\text{test}}$) are tried for the dataset `Set5' \cite{bevilacqua2012}. Experimental results are summarized in Table \ref{tab:SRCNN_Factor_Test}. 

Performance is degraded if $s_{\text{train}} \neq s_{\text{test}}$. For scale factor 2, the model trained with factor 2 gives PSNR of 37.10 (in dB), whereas models trained with factor 3 and 4 give 30.05 and 28.13, respectively. A network trained over single-scale data is not capable of handling other scales. In many tests, it is even worse than bicubic interpolation, the method used for generating the input image. 

We now test if a model trained with scale augmentation is capable of performing SR at multiple scale factors. The same network used above is trained with multiple scale factors $s_{\text{train}} = \{2,3,4\}$. In addition, we experiment with the cases $s_{\text{train}} = \{2,3\}, \{2,4\}, \{3,4\}$ for more comparisons. 

We observe that the network copes with any scale used during training. When $s_{\text{train}} = \{2,3,4\}$ ($\times 2, 3, 4$ in Table \ref{tab:SRCNN_Factor_Test}), its PSNR for each scale is comparable to those achieved from the corresponding result of single-scale network: 37.06 vs. 37.10 ($\times 2$), 33.27 vs. 32.89 ($\times 3$), 30.95 vs. 30.86 ($\times 4$).

Another pattern is that for large scales ($\times 3,4$), our multi-scale network outperforms single-scale network: our model ($\times 2,3$), ($\times 3,4$) and ($\times 2, 3,4$) give PSNRs 33.22, 33.24 and 33.27 for test scale 3, respectively, whereas ($\times 3$) gives 32.89. Similarly, ($\times 2,4$), ($\times 3,4$) and ($\times 2, 3,4$) give 30.86, 30.94 and 30.95 (vs. 30.84 by $\times 4$ model),  respectively. From this, we observe that training multiple scales boosts the performance for large scales.

\section{Experimental Results}
\label{sec:exp}
In this section, we evaluate the performance of our method on several datasets. We first describe datasets used for training and testing our method. Next, parameters necessary for training are given. 

After outlining our experimental setup, we compare our method with several state-of-the-art SISR methods. 

\subsection{Datasets for Training and Testing}
\textbf{Training dataset} Different learning-based methods use different training images. For example, RFL \cite{schulter2015fast} has two methods, where the first one uses 91 images from Yang et al. \cite{yang2010image} and the second one uses 291 images with the addition of 200 images from Berkeley Segmentation Dataset \cite{Martin2001}. SRCNN \cite{dong2015image} uses a very large ImageNet dataset. 

We use 291 images as in \cite{schulter2015fast} for benchmark with other methods in this section. In addition, data augmentation (rotation or flip) is used. For results in previous sections, we used 91 images to train network fast, so performances can be slightly different. 

\textbf{Test dataset} For benchmark, we use four datasets. Datasets `Set5' \cite{bevilacqua2012} and `Set14' \cite{zeyde2012single} are often used for benchmark in other works \cite{Timofte,Timofte2013,Dong2014}. Dataset `Urban100', a dataset of urban images recently provided by Huang et al. \cite{Huang-CVPR-2015}, is very interesting as it contains many challenging images failed by many of the existing methods. Finally, dataset `B100', natural images in the Berkeley Segmentation Dataset used in Timofte et al. \cite{Timofte} and Yang and Yang \cite{Yang2013} for benchmark, is also employed. 
\subsection{Training Parameters}
We provide parameters used to train our final model. We use a network of depth 20. Training uses batches of size 64. Momentum and weight decay parameters are set to 0.9 and $0.0001$, respectively. 

For weight initialization, we use the method described in He et al. \cite{he2015delving}. This is a theoretically sound procedure for networks utilizing rectified linear units (ReLu).

We train all experiments over 80 epochs (9960 iterations with batch size 64). Learning rate was initially set to 0.1 and then decreased by a factor of 10 every 20 epochs. In total, the learning rate was decreased 3 times, and the learning
is stopped after 80 epochs. Training takes roughly 4 hours on GPU Titan Z. 

%\footnotetext{\textcolor{red}{PSNRs are slightly different from the original paper as they use different evaluation framework from Timofte et al. \cite{Timofte}}}

\subsection{Benchmark}
For benchmark, we follow the publicly available framework of Huang et al. \cite{Timofte2013}. It enables the comparison of many state-of-the-art results with the same evaluation procedure.

The framework applies bicubic interpolation to color components of an image and sophisticated models to luminance components as in  other methods \cite{chang2004super}, \cite{glasner2009super}, \cite{zeyde2012single}. This is because human vision is more sensitive to details in intensity than in color.

This framework crops pixels near image boundary. For our method, this procedure is unnecessary as our network outputs the full-sized image. For fair comparison, however, we also crop pixels to the same amount.

\subsection{Comparisons with State-of-the-Art Methods}
We provide quantitative and qualitative comparisons. Compared methods are A+ \cite{Timofte}, RFL\cite{schulter2015fast}, SelfEx \cite{Huang-CVPR-2015} and SRCNN \cite{Dong2014}. In Table \ref{tbl:benchmark}, we provide a summary of quantitative evaluation on several datasets. Our methods outperform all previous methods in these datasets. Moreover, our methods are relatively fast. The public code of SRCNN based on a CPU implementation is slower than the code used by Dong et. al \cite{dong2015image} in their paper based on a GPU implementation.

In Figures \ref{fig:c2} and \ref{fig:c3}, we compare our method with top-performing methods. In Figure \ref{fig:c2}, only our method perfectly reconstructs the line in the middle. Similarly, in Figure \ref{fig:c3}, contours are clean and vivid in our method whereas they are severely blurred or distorted in other methods. 

\section{Conclusion}
In this work, we have presented a super-resolution method using very deep networks. Training a very deep network is hard due to a slow convergence rate. We use residual-learning and extremely high learning rates to optimize a very deep network fast. Convergence speed is maximized and we use gradient clipping to ensure the training stability. We have demonstrated that our method outperforms the existing method by a large margin on benchmarked images. We believe our approach is readily applicable to other image restoration problems such as denoising and compression artifact removal.

{\small
	\bibliographystyle{ieee}
	\bibliography{VDSR}
}
\end{document}